# An Information-Theoretic Analysis of Hard and Soft Assignment Methods for Clustering

**Michael Kearns**
AT&T Labs Research
Florham Park, New Jersey

**Yishay Mansour**
Tel Aviv University
Tel Aviv, Israel

**Andrew Y. Ng**
Carnegie Mellon University
Pittsburgh, Pennsylvania

## Abstract

Assignment methods are at the heart of many algorithms for unsupervised learning and clustering — in particular, the well-known $K$-*means* and *Expectation-Maximization* (EM) algorithms. In this work, we study several different methods of assignment, including the "hard" assignments used by $K$-means and the "soft" assignments used by EM. While it is known that $K$-means minimizes the distortion on the data and EM maximizes the likelihood, little is known about the systematic differences of behavior between the two algorithms. Here we shed light on these differences via an information-theoretic analysis. The cornerstone of our results is a simple decomposition of the expected distortion, showing that $K$-means (and its extension for inferring general parametric densities from unlabeled sample data) must implicitly manage a trade-off between how similar the data assigned to each cluster are, and how the data are *balanced* among the clusters. How well the data are balanced is measured by the entropy of the partition defined by the hard assignments. In addition to letting us predict and verify systematic differences between $K$-means and EM on specific examples, the decomposition allows us to give a rather general argument showing that $K$-means will consistently find densities with less "overlap" than EM. We also study a third natural assignment method that we call *posterior* assignment, that is close in spirit to the soft assignments of EM, but leads to a surprisingly different algorithm.

## 1 Introduction

Algorithms for density estimation, clustering and unsupervised learning are an important tool in machine learning. Two classical algorithms are the $K$-*means* algorithm [7, 1, 3] and the *Expectation-Maximization* (EM) algorithm [2]. These algorithms have been applied in a wide variety of settings, including parameter estimation in hidden Markov models for speech recognition [8], estimation of conditional probability tables in belief networks for probabilistic inference [6], and various clustering problems [3].

At a high level, $K$-means and EM appear rather similar: both perform a two-step iterative optimization, performed repeatedly until convergence. The first step is an *assignment* of data points to "clusters" or density models, and the second step is a *reestimation* of the clusters or density models based on the current assignments. The $K$-means and EM algorithms differ only in the manner in which they assign data points (the first step). Loosely speaking, in the case of two clusters [1], if $P_0$ and $P_1$ are density models for the two clusters, then $K$-means assigns $x$ to $P_0$ if and only if $P_0(x) \geq P_1(x)$; otherwise $x$ is assigned to $P_1$. We call this *hard* or *Winner-Take-All* (WTA) assignment. In contrast, EM assigns $x$ fractionally, assigning $x$ to $P_0$ with weight $P_0(x)/(P_0(x) + P_1(x))$, and assigning the "rest" of $x$ to $P_1$. We call this *soft* or *fractional* assignment. A third natural alternative would be to again assign $x$ to only one of $P_0$ and $P_1$ (as in $K$-means), but to *randomly* assign it, assigning to $P_0$ with *probability* $P_0(x)/(P_0(x) + P_1(x))$. We call this *posterior* assignment.

Each of these three assignment methods can be interpreted as classifying points as belonging to one (or more) of two distinct populations, solely on the basis of probabilistic models (densities) for these two populations. An alternative interpretation is that we have three different ways of inferring the value of a "hidden" (unobserved) variable, whose value would indicate which of two sources had generated an observed data point. How these assignment methods differ in the context of unsupervised learning is the subject of this paper.

In the context of unsupervised learning, EM is typically viewed as an algorithm for mixture density estimation. In classical density estimation, a finite training set of unlabeled data is used to derive a hypothesis density. The goal is for the hypothesis density to model the "true" sampling density as accurately as possible, typically as measured by the Kullback-Leibler (KL) divergence. The EM algorithm can be used to find a *mixture* density model of the form $\alpha_0 P_0 + (1-\alpha_0) P_1$. It is known that the mixture model found by EM will be a local minimum of the log-loss [2] (which is equivalent to a local maximum of the likelihood), the empirical analogue of the KL divergence.

---

[1]Throughout the paper, we concentrate on the case of just two clusters or densities for simplicity of development. All of our results hold for the general case of $K$ clusters or densities.

The $K$-means algorithm is often viewed as a *vector quantization* algorithm (and is sometimes referred to as the *Lloyd-Max* algorithm in the vector quantization literature). It is known that $K$-means will find a local minimum of the *distortion* or *quantization error* on the data [7], which we will discuss at some length.

Thus, for both the fractional and WTA assignment methods, there is a natural and widely used iterative optimization heuristic (EM and $K$-means, respectively), and it is known what *loss function* is (locally) minimized by each algorithm (log-loss and distortion, respectively). However, relatively little seems to be known about the precise relationship between the two loss functions and their attendant heuristics. The structural similarity of EM and $K$-means often leads to their being considered closely related or even roughly equivalent. Indeed, Duda and Hart [3] go as far as saying that $K$-means can be viewed as "an approximate way to obtain maximum likelihood estimates for the means", which is the goal of density estimation in general and EM in particular. Furthermore, $K$-means is formally equivalent to EM using a mixture of Gaussians with covariance matrices $\epsilon I$ (where $I$ is the identity matrix) in the limit $\epsilon \to 0$. In practice, there is often some conflation of the two algorithms: $K$-means is sometimes used in density estimation applications due to its more rapid convergence, or at least used to obtain "good" initial parameter values for a subsequent execution of EM.

But there are also simple examples in which $K$-means and EM converge to rather different solutions, so the preceding remarks cannot tell the entire story. What quantitative statements can be made about the systematic differences between these algorithms and loss functions?

In this work, we answer this question by giving a new interpretation of the classical distortion that is locally minimized by the $K$-means algorithm. We give a simple information-theoretic decomposition of the expected distortion that shows that $K$-means (and any other algorithm seeking to minimize the distortion) must manage a trade-off between how well the data are *balanced* or distributed among the clusters by the hard assignments, and the *accuracy* of the density models found for the two sides of this assignment. The degree to which the data are balanced among the clusters is measured by the *entropy* of the partition defined by the assignments. We refer to this trade-off as the *information-modeling* trade-off.

The information-modeling trade-off identifies two significant ways in which $K$-means and EM differ. First, where EM seeks to model the *entire* sampling density $Q$ with a *mixture* model $\alpha_0 P_0 + (1-\alpha_0) P_1$, $K$-means is concerned with explicitly identifying *distinct* subpopulations $Q_0$ and $Q_1$ of the sampling density, and finding good models $P_0$ and $P_1$ for each *separately*. Second, the choice of subpopulations identified by $K$-means may be strongly influenced by the entropy of the partition they define; in EM this influence is entirely absent. The first of these differences is the intuitive result of the differing assignment methods, and we formalize it here; the second is less obvious, but actually can determine the behavior of $K$-means even in simple examples, as we shall see.

In addition to letting us predict and explain the behavior of $K$-means on specific examples, the new decomposition allows us to derive a general prediction about how $K$-means and EM differ: namely, that $K$-means will tend to find density models $P_0$ and $P_1$ that have less "overlap" with each other compared to those found by EM. In certain simple examples, this bias of $K$-means is apparent; here we argue that it is a rather general bias that depends little on the sampling density or the form of the density models $P_0$ and $P_1$ used by the algorithms.

The mathematical framework we use also allows us to analyze the variant of $K$-means that maintains unequal weightings of the density models $P_0$ and $P_1$; we show that the use of this weighting has an interesting effect on the loss function, essentially "erasing" the incentive for finding a partition with high entropy. We also study the posterior assignment method mentioned above, and show that despite the resulting loss function's algebraic similarity to the iterative optimization performed by EM, it differs rather dramatically.

Our results should be of some interest to anyone applying EM, $K$-means and their variants to problems of unsupervised learning.

## 2    A Loss Decomposition for Hard Assignments

Suppose that we have densities $P_0$ and $P_1$ over $X$, and a (possibly randomized) mapping $F$ that maps $x \in X$ to either 0 or 1; we will refer to $F$ as a *partition* of $X$. We think of $F$ as "assigning" points to exactly one of $P_0$ and $P_1$, and we think of $P_b$ ($b \in \{0,1\}$) as a density model for the points assigned to it. $F$ may flip coins to determine the assignment of $x$, but must always output a value in $\{0,1\}$; in other words, $F$ must make "hard" assignments. We will call such a triple $(F, \{P_0, P_1\})$ a *partitioned density*. In this section, we propose a measure of goodness for partitioned densities and explore its interpretation and consequences.

In all of the settings we consider in this paper, the partition $F$ will actually be *determined* by $P_0$ and $P_1$ (and perhaps some additional parameters), but we will suppress the dependency of $F$ on these quantities for notational brevity. As simple examples of such hard assignment methods, we have the two methods discussed in the introduction: *WTA* assignment (used by $K$-means), in which $x$ is assigned to $P_0$ if and only if $P_0(x) \geq P_1(x)$, and what we call *posterior* assignment, in which $x$ is assigned to $P_b$ with probability $P_b(x)/(P_0(x) + P_1(x))$. The soft or fractional assignment method used by EM does not fall into this framework, since $x$ is fractionally assigned to *both* $P_0$ and $P_1$.

Throughout the development, we will assume that unclassified data is drawn according to some fixed, unknown density or distribution $Q$ over $X$ that we will call the *sampling* density. Now given a partitioned density $(F, \{P_0, P_1\})$, what is a reasonable way to measure how well the partitioned density "models" the sampling density $Q$? As far as the $P_b$ are concerned, as we have mentioned, we might ask that the density $P_b$ be a good model of the sampling density $Q$ *conditioned* on the event $F(x) = b$. In other words, we imagine that $F$ partitions $Q$ into two distinct subpopulations, and demand that $P_0$ and $P_1$ *separately* model these subpopulations. It is not immediately clear what criteria (if any) we should ask $F$ to meet; let us defer this question for a moment.

Fix any partitioned density $(F, \{P_0, P_1\})$, and define for any $x \in X$ the *partition loss*

$$\chi(x) = \mathbf{E}\left[-\log(P_{F(x)}(x))\right] \quad (1)$$

where the expectation is only over the (possible) randomization in $F$. We have suppressed the dependence of $\chi$ on the partitioned density under consideration for notational brevity, and the logarithm is base 2. If we ask that the partition loss be minimized, we capture the informal measure of goodness proposed above: we first use the assignment method $F$ to assign $x$ to either $P_0$ or $P_1$; and we then "penalize" only the *assigned* density $P_b$ by the log loss $-\log(P_b(x))$. We can define the *training* partition loss on a finite set of points $S$, and the *expected* partition loss with respect to $Q$, in the natural ways.

Let us digress briefly here to show that in the special case that $P_0$ and $P_1$ are multivariate Gaussian (normal) densities with means $\mu_0$ and $\mu_1$, and identity covariance matrices, and the partition $F$ is the WTA assignment method, then the partition loss on a set of points is equivalent to the well-known *distortion* or *quantization error* of $\mu_0$ and $\mu_1$ on that set of points (modulo some additive and multiplicative constants). The distortion of $x$ with respect to $\mu_0$ and $\mu_1$ is simply $(1/2)\min(||x-\mu_0||^2, ||x-\mu_1||^2) = (1/2)||x-\mu_{F(x)}||^2$, where $F(x)$ assigns $x$ to the nearer of $\mu_0$ and $\mu_1$ according to Euclidean distance (WTA assignment). Now for any $x$, if $P_b$ is the $d$-dimensional Gaussian $(1/(2\pi)^{(d/2)})e^{-(1/2)||x-\mu_b||^2}$ and $F$ is WTA assignment with respect to the $P_b$, then the partition loss on $x$ is

$$\begin{aligned}
-\log(P_{F(x)}(x)) &= \log\left((2\pi)^{d/2} e^{(1/2)||x-\mu_{F(x)}||^2}\right) \\
&= (1/2)||x-\mu_{F(x)}||^2 \log(e) \\
&\quad + (d/2)\log 2\pi. \quad (3)
\end{aligned}$$

The first term in Equation (3) is the distortion times a constant, and the second term is an additive constant that does not depend on $x$, $P_0$ or $P_1$. Thus, minimization of the partition loss is equivalent to minimization of the distortion. More generally, if $x$ and $\mu$ are equal dimensioned real vectors, and if we measure distortion using any distance metric $d(x, \mu)$ that can be expressed as a function of $x - \mu$, (that is, the distortion on $x$ is the smaller of the two distances $d(x, \mu_0)$ and $d(x, \mu_1)$,) then again this distortion is the special case of the partition loss in which the density $P_b$ is $P_b(x) = (1/Z)e^{-d(x, \mu_b)}$, and $F$ is WTA assignment. The property that $d(x, \mu)$ is a function of $x - \mu$ is a sufficient condition to ensure that the normalization factor $Z$ is independent of $\mu$; if $Z$ depends on $\mu$, then the partition loss will include an additional $\mu$-dependent term besides the distortion, and we cannot guarantee in general that the two minimizations are equivalent.

Returning to the development, it turns out that the expectation of the partition loss with respect to the sampling density $Q$ has an interesting decomposition and interpretation. For this step we shall require some basic but important definitions. For any fixed mapping $F$ and any value $b \in \{0, 1\}$, let us define $w_b = \mathbf{Pr}_{x \in Q}[F(x) = b]$, so $w_0 + w_1 = 1$. Then we define $Q_b$ by

$$Q_b(x) = Q(x) \cdot \mathbf{Pr}[F(x) = b]/w_b \quad (4)$$

where here the probability is taken only over any randomization of the mapping $F$. Thus, $Q_b$ is simply the distribution $Q$ conditioned on the event $F(x) = b$, so $F$ "splits" $Q$ into $Q_0$ and $Q_1$: that is, $Q(x) = w_0 Q_0(x) + w_1 Q_1(x)$ for all $x$. Note that *the definitions of $w_b$ and $Q_b$ depend on the partition $F$* (and therefore on the $P_b$, when $F$ is determined by the $P_b$).

Now we can write the expectation of the partition loss with respect to $Q$:

$$\begin{aligned}
\mathbf{E}_{x \in Q}[\chi(x)] &= w_0 \mathbf{E}_{x_0 \in Q_0}[-\log(P_0(x_0))] \\
&\quad + w_1 \mathbf{E}_{x_1 \in Q_1}[-\log(P_1(x_1))] \quad (5) \\
&= w_0 \mathbf{E}_{x_0 \in Q_0}\left[\log \frac{Q_0(x_0)}{P_0(x_0)} - \log(Q_0(x_0))\right] \\
&\quad + w_1 \mathbf{E}_{x_1 \in Q_1}\left[\log \frac{Q_1(x_1)}{P_1(x_1)} - \log(Q_1(x_1))\right] \quad (6) \\
&= w_0 KL(Q_0 || P_0) + w_1 KL(Q_1 || P_1) \\
&\quad + w_0 \mathcal{H}(Q_0) + w_1 \mathcal{H}(Q_1) \quad (7) \\
&= w_0 KL(Q_0 || P_0) + w_1 KL(Q_1 || P_1) \\
&\quad + \mathcal{H}(Q|F). \quad (8)
\end{aligned}$$

Here $KL(Q_b || P_b)$ denotes the Kullback-Leibler divergence from $Q_b$ to $P_b$, and $\mathcal{H}(Q|F)$ denotes $\mathcal{H}(x|F(x))$,

the entropy of the random variable $x$, distributed according to $Q$, when we are given its (possibly randomized) assignment $F(x)$.

This decomposition will form the cornerstone of all of our subsequent arguments, so let us take a moment to examine and interpret it in some detail. First, let us remember that every term in Equation (8) depends on all of $F$, $P_0$ and $P_1$, since $F$ and the $P_b$ are themselves coupled in a way that depends on the assignment method. With that caveat, note that the quantity $KL(Q_b||P_b)$ is the natural measure of how well $P_b$ models its respective side of the partition defined by $F$, as discussed informally above. Furthermore, the weighting of these terms in Equation (8) is the natural one. For instance, as $w_0$ approaches 0 (and thus, $w_1$ approaches 1), it becomes less important to make $KL(Q_0||P_0)$ small: if the partition $F$ assigns only a negligible fraction of the population to category 0, it is not important to model that category especially well, but very important to accurately model the dominant category 1. In isolation, the terms $w_0 KL(Q_0||P_0) + w_1 KL(Q_1||P_1)$ encourage us to choose $P_b$ such that the two sides of the split of $Q$ defined by $P_0$ and $P_1$ (that is, by $F$) are in fact modeled well by $P_0$ and $P_1$. But these terms are not in isolation.

The term $\mathcal{H}(Q|F)$ in Equation (8) measures the *informativeness* of the partition $F$ defined by $P_0$ and $P_1$, that is, how much it reduces the entropy of $Q$. More precisely, by appealing to the symmetry of the mutual information $\mathcal{I}(x, F(x))$, we may write (where $x$ is distributed according to $Q$):

$$\begin{align}
\mathcal{H}(Q|F) &= \mathcal{H}(x|F(x)) \tag{9}\\
&= \mathcal{H}(x) - \mathcal{I}(x, F(x)) \tag{10}\\
&= \mathcal{H}(x) - (\mathcal{H}(F(x)) - \mathcal{H}(F(x)|x)) \tag{11}\\
&= \mathcal{H}(x) - (\mathcal{H}_2(w_0) - \mathcal{H}(F(x)|x)) \tag{12}
\end{align}$$

where $\mathcal{H}_2(p) = -p\log(p) - (1-p)\log(1-p)$ is the binary entropy function. The term $\mathcal{H}(x) = \mathcal{H}(Q)$ is independent of the partition $F$. Thus, we see from Equation (12) that $F$ *reduces* the uncertainty about $x$ by the amount $\mathcal{H}_2(w_0) - \mathcal{H}(F(x)|x)$. Note that if $F$ is a deterministic mapping (as in WTA assignment), then $\mathcal{H}(F(x)|x) = 0$, and a good $F$ is simply one that maximizes $\mathcal{H}(w_0)$. In particular, *any* deterministic $F$ such that $w_0 = 1/2$ is optimal in this respect, regardless of the resulting $Q_0$ and $Q_1$. In the general case, $\mathcal{H}(F(x)|x)$ is a measure of the randomness in $F$, and a good $F$ must trade off between the competing quantities $\mathcal{H}_2(w_0)$ (which, for example, is *maximized* by the $F$ that flips a coin on every $x$) and $-\mathcal{H}(F(x)|x)$ (which is always *minimized* by this same $F$).

Perhaps most important, we expect that there may be competition between the *modeling* terms $w_0 KL(Q_0||P_0) + w_1 KL(Q_1||P_1)$ and the *partition information* term $\mathcal{H}(Q|F)$. If $P_0$ and $P_1$ are chosen from some parametric class $\mathcal{P}$ of densities of limited complexity (for instance, multivariate Gaussian distributions), then the demand that the $KL(Q_b||P_b)$ be small can be interpreted as a demand that the partition $F$ yield $Q_b$ that are "simple"(by virtue of their being well-approximated, in the KL divergence sense, by densities lying in $\mathcal{P}$). This demand may be in tension with the demand that $F$ be informative, and Equation (8) is a prescription for how to manage this competition, which we refer to in the sequel as the *information-modeling trade-off*.

Thus, if we view $P_0$ and $P_1$ as implicitly defining a hard partition (as in the case of WTA assignment), then the partition loss provides us with one particular way of evaluating the goodness of $P_0$ and $P_1$ as models of the sampling density $Q$. Of course, there are other ways of evaluating the $P_b$, one of them being to evaluate the *mixture* $(1/2)P_0 + (1/2)P_1$ via the KL divergence $KL(Q||(1/2)P_0 + (1/2)P_1)$ (we will discuss the more general case of nonequal mixture coefficients shortly). This is the expression that is (locally) minimized by standard density estimation approaches such as EM, and we would particularly like to call attention to the ways in which Equation (8) differs from this expression. Not only does Equation (8) differ by incorporating the penalty $\mathcal{H}(Q|F)$ for the partition $F$, but instead of asking that the mixture $(1/2)P_0 + (1/2)P_1$ model the entire population $Q$, each $P_b$ is only asked to — and only given credit for — modeling its respective $Q_b$. We will return to these differences in considerably more detail in Section 4.

We close this section by observing that if $P_0$ and $P_1$ are chosen from a class $\mathcal{P}$ of densities, and we constrain $F$ to be the WTA assignment method for the $P_b$, there is a simple and perhaps familiar iterative optimization algorithm for locally minimizing the partition loss on a set of points $S$ over all choices of the $P_b$ from $\mathcal{P}$ — we simply repeat the following two steps until convergence:

- (WTA Assignment) Set $S_0$ to be the set of points $x \in S$ such that $P_0(x) \geq P_1(x)$, and set $S_1$ to be $S - S_0$.
- (Reestimation) Replace each $P_b$ with $argmin_{P \in \mathcal{P}}\{-\sum_{x \in S_b} \log(P(x))\}$.

As we have already noted, in the case that the $P_b$ are restricted to be Gaussian densities with identity covariance matrices (and thus, only the means are parameters), this algorithm reduces to the classical $K$-means algorithm. Here we have given a natural extension for estimating $P_0$ and $P_1$ from a general parametric class, so we may have more parameters than just the means. With some abuse of terminology, we will simply refer to our generalized version as $K$-means. The reader familiar with the EM algorithm for choosing $P_0$ and $P_1$ from $\mathcal{P}$ will also recognize this algorithm as simply a "hard" or WTA assignment variant of *unweighted* EM (that is, where the mixture coefficients must be equal).

It is easy to verify that $K$-means will result in a local minimum of the partition loss over $P_b$ chosen from $\mathcal{P}$

using the WTA assignment method. Let us rename this special case of the partition loss the *K-means loss* for convenience.

The fact that $K$-means locally minimizes the $K$-means loss, combined with Equation (8), implies that $K$-means must implicitly manage the information-modeling trade-off. Note that although $K$-means will not increase the $K$-means loss at any iteration, this does *not* mean that each of the terms in Equation (8) will not increase; indeed, we will see examples where this is not the case. It has been often observed in the vector quantization literature [4] that at each iteration, the means estimated by $K$-means must in fact be the true means of the points assigned to them — but this does not imply, for instance, that the terms $KL(Q_b||P_b)$ are nonincreasing (because, for example, $Q_b$ can also change with each iteration).

Finally, note that we can easily generalize Equation (8) to the $K$-cluster case:

$$\mathbf{E}_Q[\chi(x)] = \sum_{i=1}^{K} w_i KL(Q_i||P_i) + \mathcal{H}(Q|F). \quad (13)$$

Note that, as in Equation (11), $\mathcal{H}(Q|F) = \mathcal{H}(x) - (\mathcal{H}(F(x)) - \mathcal{H}(F(x)|x))$, where $x$ is distributed according to $Q$, and that for general $K$, $\mathcal{H}(F(x))$ is now an $O(\log(K))$ quantity.

## 3  Weighted $K$-Means

As we have noted, $K$-means is a hard-assignment variant of the *unweighted* EM algorithm (that is, where the mixture coefficients are forced to be $1/2$, or $1/K$ in the general case of $K$ densities). There is also a natural generalization of $K$-means that can be thought of as a hard-assignment variant of *weighted* EM. For any class $\mathcal{P}$ of densities over a space $X$, *weighted $K$-means* over $\mathcal{P}$ takes as input a set $S$ of data points and outputs a pair of densities $P_0, P_1 \in \mathcal{P}$, as well as a weight $\alpha_0 \in [0, 1]$. (Again, the generalization to the case of $K$ densities and $K$ weights is straightforward.) The algorithm begins with random choices for the $P_b \in \mathcal{P}$ and $\alpha_0$, and then repeatedly executes the following three steps:

- (WTA Assignment) Set $S_0$ to be the set of points $x \in S$ such that $\alpha_0 P_0(x) \geq (1 - \alpha_0) P_1(x)$, and set $S_1$ to be $S - S_0$.
- (Reestimation) Replace each $P_b$ with $argmin_{P \in \mathcal{P}} \{-\sum_{x \in S_b} \log(P(x))\}$.
- (Reweighting) Replace $\alpha_0$ with $|S_0|/|S|$.

Now we can again ask the question: what loss function is this algorithm (locally) minimizing? Let us fix $F$ to be the *weighted WTA partition*, given by $F(x) = 0$ if and only if $\alpha_0 P_0(x) \geq (1 - \alpha_0) P_1(x)$. Note that $F$ is deterministic, and also that in general, $\alpha_0$ (which is an adjustable parameter of the weighted $K$-means algorithm) is *not* necessarily the same as $w_0$ (which is defined by the current weighted WTA partition, and depends on $Q$).

It turns out that weighted $K$-means will *not* find $P_0$ and $P_1$ that give a local minimum of the unweighted $K$-means loss, but of a slightly different loss function whose expectation differs from that of the unweighted $K$-means loss in an interesting way. Let us define the *weighted $K$-means* loss of $P_0$ and $P_1$ on $x$ by

$$-\log\left(\alpha_0^{1-F(x)}(1-\alpha_0)^{F(x)} P_{F(x)}(x)\right) \quad (14)$$

where again, $F$ is the weighted WTA partition determined by $P_0$, $P_1$ and $\alpha_0$. For any data set $S$, define $S_b = \{x \in S : F(x) = b\}$. We now show that weighted $K$-means will in fact not increase the weighted $K$-means loss on $S$ with each iteration. Thus[2]

$$-\sum_{x \in S} \log\left(\alpha_0^{1-F(x)}(1-\alpha_0)^{F(x)} P_{F(x)}(x)\right)$$

$$= -\sum_{x \in S_0} \log(\alpha_0 P_0(x))$$
$$\quad - \sum_{x \in S_1} \log((1-\alpha_0) P_1(x)) \quad (15)$$

$$= -\sum_{x \in S_0} \log(P_0(x)) - \sum_{x \in S_1} \log(P_1(x))$$
$$\quad - |S_0|\log(\alpha_0) - |S_1|\log(1-\alpha_0). \quad (16)$$

Now

$$-|S_0|\log(\alpha_0) - |S_1|\log(1-\alpha_0)$$
$$= -|S|\left(\frac{|S_0|}{|S|}\log(\alpha_0) + \frac{|S_1|}{|S|}\log(1-\alpha_0)\right) (17)$$

which is an entropic expression minimized by the choice $\alpha_0 = |S_0|/|S|$. But this is exactly the new value of $\alpha_0$ computed by weighted $K$-means from the current assignments $S_0, S_1$. Furthermore, the two summations in Equation (16) are clearly reduced by reestimating $P_0$ from $S_0$ and $P_1$ from $S_1$ to obtain the densities $P'_0$ and $P'_1$ that minimize the log-loss over $S_0$ and $S_1$ respectively, and these are again exactly the new densities computed by weighted $K$-means. Thus, weighted $K$-means decreases the weighted $K$-means loss (given by Equation (14) of $(F, \{P_0, P_1\})$ on $S$ at each iteration, justifying our naming of this loss.

Now for a fixed $P_0$ and $P_1$, what is the expected weighted $K$-means loss with respect to the sampling density $Q$? We have

$$\mathbf{E}_{x \in Q}\left[-\log\left(\alpha_0^{1-F(x)}(1-\alpha_0)^{F(x)} P_{F(x)}(x)\right)\right]$$
$$= \mathbf{E}_{x \in Q}\left[-\log(P_{F(x)}(x))\right]$$
$$\quad - w_0 \log(\alpha_0) - w_1 \log(1-\alpha_0) \quad (18)$$

---

[2] We are grateful to Nir Friedman for pointing out this derivation to us.

where $w_b = \mathbf{Pr}_{x \in X}[F(x) = b]$ as before. The first term on the right-hand side is just the expected partition loss of $(F, \{P_0, P_1\})$. The last two terms give the cross-entropy between the binary distributions $(w_0, w_1) = (w_0, 1 - w_0)$ and $(\alpha_0, 1 - \alpha_0)$. For a fixed $(F, \{P_0, P_1\})$, there is not much we can say about this cross-entropy; but for weighted $K$-means, we know that at convergence we must have $\alpha_0 = |S_0|/|S|$ (for this is how weighted $K$-means reassigns $\alpha_0$ at each iteration), and $|S_0|/|S| = \hat{w}_0$ is simply the empirical estimate of $w_0$. Thus, in the limit of large samples we expect $\hat{w}_0 \to w_0$, and thus

$$-w_0 \log(\hat{w}_0) - w_1 \log(\hat{w}_1) \to \mathcal{H}_2(w_0). \qquad (19)$$

Combining Equation (19) with Equation (18) and our general decomposition for partition loss in Equation (8) gives that *for the $P_0$, $P_1$ and $\alpha_0$ found by weighted $K$-means,*

$$\begin{aligned}
& \mathbf{E}_{x \in Q}\left[-\log\left(\alpha_0^{1-F(x)}(1-\alpha_0)^{F(x)} P_{F(x)}(x)\right)\right] \\
&= w_0 KL(Q_0\|P_0) + w_1 KL(Q_1\|P_1) + \mathcal{H}(Q|F) \\
&\quad -w_0 \log(\hat{w}_0) - w_1 \log(\hat{w}_1) \qquad (20) \\
&= w_0 KL(Q_0\|P_0) + w_1 KL(Q_1\|P_1) + \mathcal{H}(Q) - \mathcal{H}_2(w_0) \\
&\quad -w_0 \log(\hat{w}_0) - w_1 \log(\hat{w}_1) \qquad (21) \\
&\approx w_0 KL(Q_0\|P_0) + w_1 KL(Q_1\|P_1) + \mathcal{H}(Q). \qquad (22)
\end{aligned}$$

Thus, since $\mathcal{H}(Q)$ does not depend on the $P_b$ or $\alpha_0$, we may think of the (generalization) goal of weighted $K$-means as finding $(F, \{P_0, P_1\})$ that minimizes the sum $w_0 KL(Q_0\|P_0) + w_1 KL(Q_1\|P_1)$. This differs from the goal of unweighted $K$-means in two ways. First of all, the introduction of the weight $\alpha_0$ has changed our definition of the partition $F$, and thus has changed the definition of $Q_0$ and $Q_1$, even for fixed $P_0, P_1$ (unweighted $K$-means corresponds to fixing $\alpha_0 = 1/2$). But beyond this, the introduction of the weight $\alpha_0$ has also removed the bias towards finding an "informative" partition $F$. Thus, *there is no information-modeling trade-off for weighted $K$-means*; the algorithm will try to minimize the modeling terms $w_0 KL(Q_0\|P_0) + w_1 KL(Q_1\|P_1)$ only. Note, however, that this is still quite different from the *mixture* KL divergence minimized by EM.

## 4   $K$-Means vs. EM: Examples

In this section, we consider several different sampling densities $Q$, and compare the solutions found by $K$-means (both unweighted and weighted) and EM. In each example, there will be significant differences between the error surfaces defined over the parameter space by the $K$-means losses and the KL divergence. Our main tool for understanding these differences will be the loss decompositions given for the unweighted $K$-means loss by Equation (8) and for the weighted $K$-means loss by Equation (22). It is important to remember that the solutions found by one of the algorithms should not be considered "better" than those found by the other algorithms: we simply have different loss functions, each justifiable on its own terms, and the choice of which loss function to minimize (that is, which algorithm to use) determines which solution we will find.

Throughout the following examples, the instance space $X$ is simply $\Re$. We compare the solutions found by (unweighted and weighted) EM and (unweighted and weighted) $K$-means when the output is a pair $\{P_0, P_1\}$ of Gaussians over $\Re$ — thus $P_0 = \mathcal{N}(\mu_0, \sigma_0)$ and $P_1 = \mathcal{N}(\mu_1, \sigma_1)$, where $\mu_0, \sigma_0, \mu_1, \sigma_1 \in \Re$ are the parameters to be adjusted by the algorithms. (The weighted versions of both algorithms also output the weight parameter $\alpha_0 \in [0, 1]$.) In the case of EM, the output is interpreted as representing a *mixture* distribution, which is evaluated by its KL divergence from the sampling density. In the case of (unweighted or weighted) $K$-means, the output is interpreted as a partitioned density, which is evaluated by the expected (unweighted or weighted) $K$-means loss with respect to the sampling density. Note that the generalization here over the classical vector quantization case is simply in allowing the Gaussians to have non-unit variance.

In each example, the various algorithms were run on 10 thousand examples from the sampling density; for these 1-dimensional problems, this sample size is sufficient to ensure that the observed behavior is close to what it would be running directly on the sampling density.

**Example (A).** Let the sampling density $Q$ be the symmetric Gaussian mixture

$$Q = 0.5\mathcal{N}(-2, 1.5) + 0.5\mathcal{N}(2, 1.5). \qquad (23)$$

See Figure 1. Suppose we initialized the parameters for the algorithms as $\mu_0 = -2$, $\mu_1 = 2$, and $\sigma_0 = \sigma_1 = 1.5$. Thus, each algorithm begins its search from the "true" parameter values of the sampling density. The behavior of unweighted EM is clear: we are starting EM at the global minimum of its expected loss function, the KL divergence; by staying where it begins, EM can enjoy a solution that perfectly models the sampling density $Q$ (that is, KL divergence 0). The same is also true of weighted EM: the presence or absence of the weighting parameter $\alpha_0$ is essentially irrelevant here, since the optimal value for this parameter is $\alpha_0 = 0.5$ for this choice of $Q$.

What about unweighted $K$-means? Let us examine each of the terms in the decomposition of the expected partition loss given in Equation (8). The term $\mathcal{H}(Q|F)$ is already minimized by the initial choice of parameters: the WTA partition $F$ is simply $F(x) = 0$ if and only if $x \leq 0$, which yields $w_0 = 1/2$ and $\mathcal{H}_2(w_0) = 1$. The terms $w_0 KL(Q_0\|P_0)$ and $w_1 KL(Q_1\|P_1)$, however, are a different story. Notice that $Q_0$ — which is $Q$ conditioned on the event $F(x) = 0$, or $x \leq 0$ — is *not* $\mathcal{N}(-2, 1.5)$. Rather, it is $\mathcal{N}(-2, 1.5)$ "chopped off" above $x = 0$, but with the tail of $\mathcal{N}(2, 1.5)$ be-low $x = 0$ added on. Equivalently, it is $\mathcal{N}(-2, 1.5)$ with its tail above $x = 0$ reflected back below $x = 0$. Clearly, the tail reflection operation on $\mathcal{N}(-2, 1.5)$ that results in $Q_0$ moves the mean of $Q_0$ *left* of $-2$ (since the tail reflection moved mass left), and *reduces* the variance below 1.5 (since the tail has moved towards the final mean). Thus, with respect to only the term $w_0 KL(Q_0||P_0)$, the best choice of $\mu_0$ should be smaller than the initial value of $-2$, and the best choice of $\sigma_0$ should be smaller than the initial value of 1.5. Symmetric remarks apply to the term $w_1 KL(Q_1||P_1)$. Furthermore, as long as the movements of $\mu_0$ and $\mu_1$, and $\sigma_0$ and $\sigma_1$, are *symmetric*, then the WTA partition $F$ will remain *unchanged* by these movements — thus, it is possible to improve the terms $w_b KL(Q_b||P_b)$ from the initial conditions *without* degrading the initially optimal value for the term $\mathcal{H}(Q|F)$. We make essentially the same prediction for weighted $K$-means, as the optimal performance is achieved for $\alpha_0 = 0.5$.

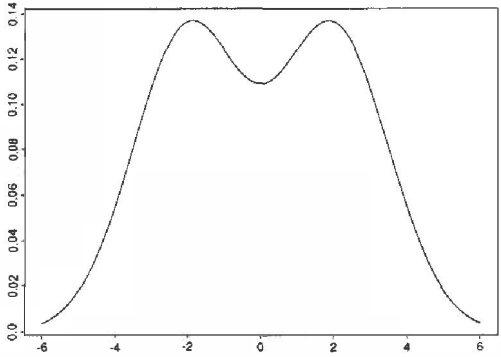

Figure 1: The sampling density for Example (A).

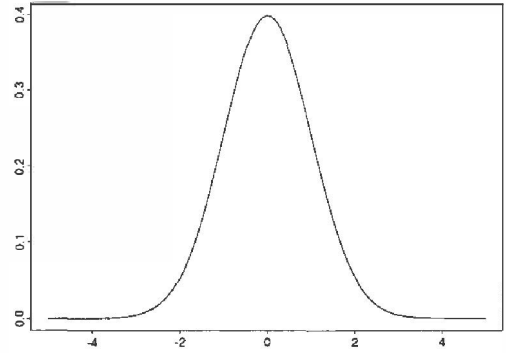

Figure 2: The sampling density for Example (B).

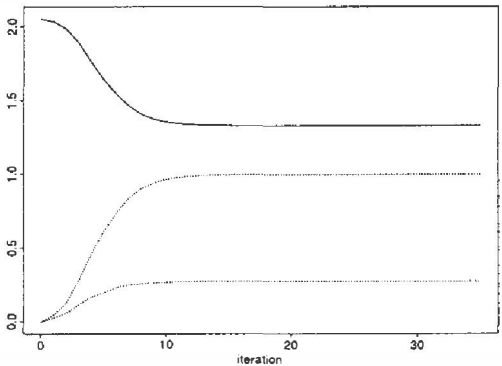

Figure 3: Evolution of the $K$-means loss (top plot) and its decomposition for Example (B): KL divergences $w_0 KL(Q_0||P_0) + w_1 KL(Q_1||P_1)$ (bottom plot) and partition information gain $\mathcal{H}_2(w_0)$ (middle plot), as a function of the iteration of unweighted $K$-means running on 10 thousand examples from $Q = \mathcal{N}(0,1)$.

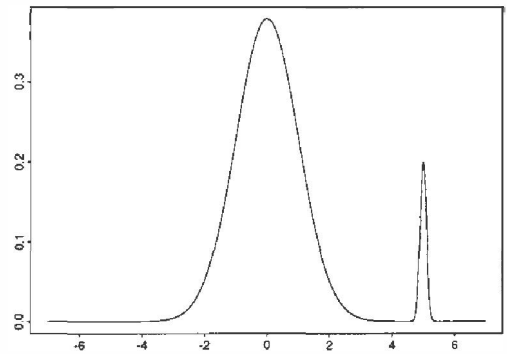

Figure 4: Plot of the sampling mixture density $Q = 0.95\mathcal{N}(0,1) + 0.05\mathcal{N}(5,0.1)$ for Example (C).

Performing the experiment on the finite sample, we find that after 8 iterations, $K$-means has converged to the solution

$$\mu_0 = -2.130, \sigma_0 = 1.338, \mu_1 = 2.131, \sigma_1 = 1.301 \quad (24)$$

which yields $w_0 = 0.500$. As predicted, the means have been pushed out from the origin, and the variances reduced. Naturally, the KL divergence from the sampling density $Q$ to the *mixture model* is inferior to that of the starting parameters, while its expected $K$-means loss is superior.

Let us remark that in this simple example, it would have been easy to predict the behavior of $K$-means directly. The point is that the decomposition of Equation (8) provides a *justification* of this behavior that cannot be provided by regarding $K$-means as a coarse approximation to EM. We now move on to some examples where the behavior of the various algorithms is more subtle.

**Example (B).** We now examine an example in which the term $\mathcal{H}(Q|F)$ directly *competes* with the KL divergences. Let the sampling density $Q$ be the single unit-variance Gaussian $Q(x) = \mathcal{N}(0, 1)$; see Figure 2. Consider the initial choice of parameters $\mu_0 = 0$, $\sigma_0 = 1$, and $P_1$ at some very distant location, say $\mu_0 = 100$, $\sigma_0 = 1$. We first examine the behavior of unweighted $K$-means. The WTA partition $F$ defined by these settings is $F(x) = 0$ if and only if $x < 50$. Since $Q$ has so little mass above $x = 50$, we have $w_0 \approx 1$, and thus $\mathcal{H}(Q|F) \approx \mathcal{H}(Q)$: the partition is not informative. The term $w_1 KL(Q_1||P_1)$ in Equation (8) is negligible, since $w_1 \approx 0$. Furthermore, $Q_0 \approx \mathcal{N}(0,1)$ because even though the tail reflection described in Example (A) occurs again here, the tail of $\mathcal{N}(0,1)$ above $x = 50$ is a negligible part of the density. Thus $w_0 KL(Q_0||P_0) \approx 0$, so $w_0 KL(Q_0||P_0) + w_1 KL(Q_1||P_1) \approx 0$. In other words, if all we cared about were the KL divergence terms, these settings would be near-optimal.

But the information-modeling trade-off is at work here: by moving $P_1$ closer to the origin, our KL divergences may degrade, but we obtain a more informative partition. Indeed, after 32 iterations unweighted $K$-means converges to

$$\mu_0 = -0.768, \sigma_0 = 0.602, \mu_1 = 0.821, \sigma_1 = 0.601 \quad (25)$$

which yields $w_0 = 0.509$.

The information-modeling tradeoff is illustrated nicely by Figure 3, where we simultaneously plot the unweighted $K$-means loss and the terms $w_0 KL(Q_0||P_0) + w_1 KL(Q_1||P_1)$ and $\mathcal{H}_2(w_0)$ as a function of the number of iterations during the run. The plot clearly shows the increase in $\mathcal{H}_2(w_0)$ (meaning a decrease in $\mathcal{H}(Q|F)$), and an increase in $w_0 KL(Q_0||P_0) + w_1 KL(Q_1||P_1)$. The fact that the gain in partition information is worth the increase in KL divergences is shown by the resulting decrease in the unweighted $K$-means loss. Note that it would be especially difficult to justify the solution found by unweighted $K$-means from the viewpoint of density estimation.

As might be predicted from Equation (22), the behavior of *weighted* $K$-means is dramatically different for this $Q$, since this algorithm has no incentive to find an informative partition, and is only concerned with the KL divergence terms. We find that after 8 iterations it has converged to

$$\mu_0 = 0.011, \sigma_0 = 0.994, \mu_1 = 3.273, \sigma_1 = 0.033 \quad (26)$$

with $\alpha_0 = w_0 = 1.000$. Thus, as expected, weighted $K$-means has chosen a completely uninformative partition, in exchange for making $w_b KL(Q_b||P_b) \approx 0$. The values of $\mu_1$ and $\sigma_1$ simply reflect the fact that at convergence, $P_1$ is assigned only the few rightmost points of the 10 thousand examples.

Note that the behavior of both $K$-means algorithms is rather different from that of EM, which will prefer $P_0 = P_1 = \mathcal{N}(0,1)$ resulting in the mixture $(1/2)P_0 + (1/2)P_1 = \mathcal{N}(0,1)$. However, the solution found by weighted $K$-means is "closer" to that of EM, in the sense that weighted $K$-means effectively eliminates one of its densities and fits the sampling density with a single Gaussian.

**Example (C).** A slight modification to the sampling distribution of Example (B) results in some interesting and subtle difference of behavior for our algorithms. Let $Q$ be given by

$$Q = 0.95\mathcal{N}(0,1) + 0.05\mathcal{N}(5, 0.1). \quad (27)$$

Thus, $Q$ is essentially as in Example (B), but with addition of a small distant "spike" of density; see Figure 4.

Starting unweighted $K$-means from the initial conditions $\mu_0 = 0, \sigma_0 = 1, \mu_1 = 0, \sigma_1 = 5$ (which has $w_0 = 0.886$, $\mathcal{H}(w_0) = 0.513$ and $w_0 KL(Q_0||P_0) + w_1 KL(Q_1||P_1) = 2.601$), we obtain convergence to the solution

$$\mu_0 = -0.219, \sigma_0 = 0.470, \mu_1 = 0.906, \sigma_1 = 1.979 \quad (28)$$

which is shown in Figure 5 (and has $w_0 = 0.564$, $\mathcal{H}(w_0) = 0.988$, and $w_0 KL(Q_0||P_0) + w_1 KL(Q_1||P_1) = 2.850$). Thus, as in Example (B), unweighted $K$-means starts with a solution that is better for the KL divergences, and worse for the partition information, and elects to degrade the former in exchange for improvement in the latter. However, it is interesting to note that $\mathcal{H}(w_0) = \mathcal{H}(0.564) = 0.988$ is still bounded significantly away from 1; presumably this is because any *further* improvement to the partition information would *not* be worth the degradation of the KL divergences. In other words, this solution found is a minimum of the $K$-means loss where there is truly a *balance* of the two terms: movement of the parameters in one direction causes the loss to increase due to a decrease in the partition information, while movement of the parameters in another direction causes the loss to increase due to an increase in the modeling error.

Unlike Example (B), there is also another (local) minimum of the unweighted $K$-means loss for this sampling density, at

$$\mu_0 = 0.018, \sigma_0 = 0.997, \mu_1 = 4.992, \sigma_1 = 0.097 \quad (29)$$

with the suboptimal unweighted $K$-means loss of 1.872. This is clearly a local minimum where the KL divergence terms are being minimized, at the expense of an uninformative partition ($w_0 = 0.949$). It is also essentially the same as the solution chosen by weighted $K$-means (regardless of the initial conditions), which is easily predicted from Equation (22).

Not surprisingly, in this example weighted $K$-means converges to a solution close to that of Equation (29).

**Example (D).** Let us examine a case in which the sampling density is a mixture of *three* Gaussians:

$$Q = 0.25\mathcal{N}(-10, 1) + 0.5\mathcal{N}(0, 1) + 0.25\mathcal{N}(10, 1). \quad (30)$$

See Figure 6. Thus, there are three rather distinct subpopulations of the sampling density. If we run unweighted $K$-means on 10 thousand examples from $Q$ from the initial conditions $\mu_0 = -5$, $\mu_1 = 5$,

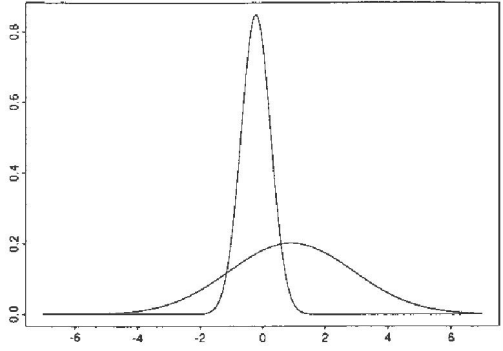

Figure 5: $P_0$ and $P_1$ found by unweighted $K$-means for the sampling density of Example (C).

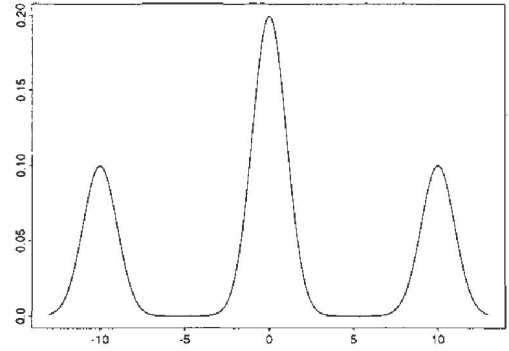

Figure 6: The sampling density for Example (D).

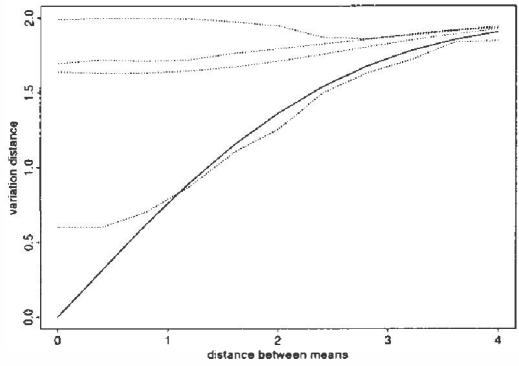

Figure 7: Variation distance $V(P_0, P_1)$ as a function of the distance between the sampling means for EM (bottom grey line), unweighted $K$-means (lowest of top three grey lines), posterior loss gradient descent (middle to top three grey lines), and weighted $K$-means (top grey line). The dark line plots $V(Q_0, Q_1)$.

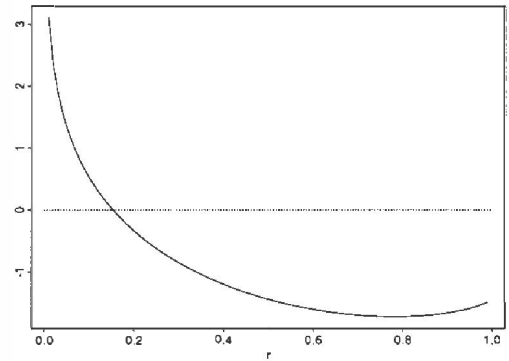

Figure 8: Plot of Equation (52) (vertical axis) as a function of $R_0 = R_0(x)$ (horizontal axis). The line $y = 0$ is also plotted as a reference.

$\sigma_0 = \sigma_1 = 1$, (which has $w_0 = 0.5$) we obtain convergence to

$$\mu_0 = -3.262, \sigma_0 = 4.789, \mu_1 = 10.006, \sigma_1 = 0.977 \quad (31)$$

which has $w_0 = 0.751$. Thus, unweighted $K$-means sacrifices the initial optimally informative partition in exchange for better KL divergences. (Weighted $K$-means converges to approximately the same solution, as we might have predicted from the fact that even the unweighted algorithm did not choose to maximize the partition information.) Furthermore, note that it has modeled two of the subpopulations of $Q$ ($\mathcal{N}(-10, 1)$ and $\mathcal{N}(0, 1)$) using $P_0$ and modeled the other ($\mathcal{N}(10, 1)$) using $P_1$. This is natural "clustering" behavior — the algorithm prefers to group the middle subpopulation $\mathcal{N}(0, 1)$ with either the left or right subpopulation, rather than "splitting" it. In contrast, unweighted EM from the same initial conditions converges to the approximately symmetric solution

$$\mu_0 = -4.599, \sigma_0 = 5.361, \mu_1 = 4.689, \sigma_1 = 5.376. \quad (32)$$

Thus, unweighted EM chooses to *split* the middle population between $P_0$ and $P_1$. The difference between $K$-means and unweighted EM in this example is a simple illustration of the difference between the quantities $w_0 KL(Q_0 || P_0) + w_1 KL(Q_1 || P_1)$ and $KL(Q || \alpha_0 P_0 + (1 - \alpha_0) P_1)$, and shows a natural case in which the behavior of $K$-means is perhaps preferable from the clustering point of view. Interestingly, in this example the solution found by *weighted* EM is again quite close to that of $K$-means.

## 5 $K$-Means Forces Different Populations

The partition loss decomposition given by Equation (8) has given us a better understanding of the loss function being minimized by $K$-means, and allowed us to explain some of the differences between $K$-means and EM on specific, simple examples. Are there any *general* differences we can identify? In this section we give a derivation that strongly suggests a bias inherent in the $K$-means algorithm: namely, a bias towards finding component densities that are as "different" as possible, in a sense to be made precise.

Let $V(P_0, P_1)$ denote the *variation distance* [3] between the densities $P_0$ and $P_1$:

$$V(P_0, P_1) = \int_x |P_0(x) - P_1(x)| dx. \quad (33)$$

Note that $V(P_0, P_1) \leq 2$ always. Notice that due to the triangle inequality, for any partitioned density $(F, \{P_0, P_1\})$,

$$V(Q_0, Q_1) \leq V(Q_0, P_0) + V(P_0, P_1) + V(Q_1, P_1). \quad (34)$$

Let us assume without loss of generality that $w_0 = \mathbf{Pr}_{x \in Q}[F(x) = 0] \leq 1/2$. Now in the case of unweighted or weighted $K$-means (or indeed, any other case where a deterministic partition $F$ is chosen), $V(Q_0, Q_1) = 2$, so from Equation (34) we may write

$$\begin{aligned} V(P_0, P_1) &\geq 2 - V(Q_0, P_0) - V(Q_1, P_1) \quad (35) \\ &= 2 - 2(w_0 V(Q_0, P_0) + w_1 V(Q_1, P_1) \\ &\quad + ((1/2) - w_0) V(Q_0, P_0) \\ &\quad + ((1/2) - w_1) V(Q_1, P_1)) \quad (36) \\ &\geq 2 - 2(w_0 V(Q_0, P_0) + w_1 V(Q_1, P_1)) \\ &\quad - 2((1/2) - w_0) V(Q_0, P_0) \quad (37) \\ &\geq 2 - 2(w_0 V(Q_0, P_0) + w_1 V(Q_1, P_1)) \\ &\quad - 2(1 - 2w_0). \quad (38) \end{aligned}$$

Let us examine Equation (38) in some detail. First, let us assume $w_0 = 1/2$, in which case $2(1 - 2w_0) = 0$. Then Equation (38) lower bounds $V(P_0, P_1)$ by a quantity that approaches the maximum value of 2 as $V(Q_0, P_0) + V(Q_1, P_1)$ approaches 0. Thus, to the extent that $P_0$ and $P_1$ succeed in approximating $Q_0$ and $Q_1$, $P_0$ and $P_1$ must differ from each other. But the partition loss decomposition of Equation (8) includes the terms $KL(Q_b || P_b)$, which are directly encouraging $P_0$ and $P_1$ to approximate $Q_0$ and $Q_1$. It is true that we are conflating two different technical senses of approximation (variation distance KL divergence). But more rigorously, since $V(P, Q) \leq \sqrt{KL(P||Q)}$ holds for any $P$ and $Q$, and for all $x$ we have $\sqrt{x} \leq x + 1/4$, we may write

$$\begin{aligned} V(P_0, P_1) &\geq 2 - 2(w_0 KL(Q_0||P_0) + w_1 KL(Q_1||P_1) + 1/4) \\ &\quad - 2(1 - 2w_0) \quad (39) \\ &= 3/2 - 2(w_0 KL(Q_0||P_0) + w_1 KL(Q_1||P_1)) \\ &\quad - 2(1 - 2w_0). \quad (40) \end{aligned}$$

Since the expression $w_0 KL(Q_0||P_0) + w_1 KL(Q_1||P_1)$ directly appears in Equation (8), we see that $K$-means is attempting to minimize a loss function that encourages $V(P_0, P_1)$ to be large, at least in the case that the algorithm finds roughly equal weight clusters ($w_0 \approx 1/2$)

---

[3] The ensuing argument actually holds for any distance metric on densities.

— which one might expect to be the case, at least for unweighted $K$-means, since there is the entropic term $-\mathcal{H}_2(w_0)$ in Equation (12). For weighted $K$-means, this entropic term is eliminated.

In Figure 7, we show the results of a simple experiment supporting the suggestion that $K$-means tends to find densities with less overlap than EM does. In the experiment, the sampling density $Q$ was a mixture of two one-dimensional, unit-variance Gaussians with varying distance between the means (the horizontal axis). The vertical axis shows the variation distance between the two target Gaussians (dark line) as a reference, and the variation distance between $P_0$ and $P_1$ for the solutions found by EM (grey line near solid line), and for unweighted $K$-means (lowest of the top three grey lines), posterior loss gradient descent, which is discussed in the next section (middle of the top three grey lines), and weighted $K$-means (top grey line).

## 6 A New Algorithm: The Posterior Partition

The WTA assignment method is one way of making hard assignments on the basis of $P_0$ and $P_1$. But there is another natural hard assignment method — perhaps even more natural. Suppose that we *randomly* assign any fixed $x$ to $P_b$ with probability $P_b(x)/(P_0(x) + P_1(x))$. Thus, we assign $x$ to $P_b$ with the posterior probability that $x$ was generated by $P_b$ under the prior assumption that the sampling density is $(1/2)P_0 + (1/2)P_1$ (which, of course, may not be true). We call this $F$ the *posterior* partition.

One nice property of the posterior partition compared to WTA assignment is that it avoids the potential "truncation" resulting from WTA assignment mentioned in Example (A) — namely, that even when $P_0$ and $P_1$ have the same form as the true sampling mixture components, we cannot make the terms $KL(Q_b||P_b)$ zero. (Recall that this occurred when the sampling density was a Gaussian mixture, the $P_b$ were Gaussian, but WTA assignment resulted in $Q_b$ that were each Gaussian with one tail "reflected back.") But if $F$ is the posterior partition, and $Q = (1/2)\tilde{Q}_0 + (1/2)\tilde{Q}_1$, and $P_0 = \tilde{Q}_0$, $P_1 = \tilde{Q}_1$ then

$$\begin{aligned} Q_b(x) &= Q(x) \cdot \mathbf{Pr}[F(x) = b]/w_b \quad (41) \\ &= (\tilde{Q}_0(x) + \tilde{Q}_1(x)) \left( \frac{\tilde{Q}_b(x)}{\tilde{Q}_0(x) + \tilde{Q}_1(x)} \right) (42) \\ &= \tilde{Q}_b(x) \quad (43) \\ &= P_b(x). \quad (44) \end{aligned}$$

If we have $Q = (1/2)\tilde{Q}_0 + (1/2)\tilde{Q}_1$, and $P_0 = \tilde{Q}_0$, $P_1 = \tilde{Q}_1$, then by the above derivation $w_b KL(Q_b||P_b) = 0$. Thus, the KL divergence terms in the expected partition loss given by Equation (8) encourage us to model the sampling density under this definition of $F$. For this reason, it is tempting to think that the use of the posterior partition will lead us closer to density estimation than will WTA assignments. However, the situation is more subtle than this, again because of the competing constraint for an informative partition. We will see an example in a moment.

Note that under the posterior partition $F$, the partition loss of $(F, \{P_0, P_1\})$ on a fixed point $x$ is

$$\begin{aligned}
\mathbf{E}[\chi(x)] &= \mathbf{E}\left[-\log P_{F(x)}(x)\right] \\
&= -\frac{P_0(x)}{P_0(x) + P_1(x)} \log P_0(x) \\
&\quad -\frac{P_1(x)}{P_0(x) + P_1(x)} \log P_1(x) \quad (45)
\end{aligned}$$

where here the expectation is taken over only the randomization of $F$; we will call this special case of the partition loss the *posterior loss*. The posterior loss on a sample $S$ is then simply the summation of the right-hand-side of Equation (45) over all $x \in S$.

**Example (A) Revisited.** Recall that the sampling density in Example (A) is

$$Q = 0.5\mathcal{N}(-2, 1.5) + 0.5\mathcal{N}(2, 1.5) \quad (46)$$

and that if we start at $P_0 = \tilde{Q}_0 = \mathcal{N}(-2, 1.5)$, $P_1 = \tilde{Q}_1 = \mathcal{N}(2, 1.5)$, then $K$-means (both weighted and unweighted) will move the means away from the origin symmetrically, since a maximally informative partition $F$ is preserved by doing so, and the KL divergences are improved. Under the posterior partition definition of $F$, the KL divergences *cannot* be improved from these initial conditions — but the informativeness of the partition can! This is because our general expression for $\mathcal{H}(x|F(x))$ is $\mathcal{H}(x) - (\mathcal{H}_2(w_0) - \mathcal{H}(F(x)|x))$ (here $x$ is distributed according to $Q$). In the $K$-means choice of $F$, the term $\mathcal{H}(F(x)|x))$ was 0, as $F$ was deterministic. Under the posterior partition, at the stated initial conditions $\mathcal{H}_2(w_0) = \mathcal{H}_2(1/2) = 1$ still holds, but now $\mathcal{H}(F(x)|x)) \neq 0$, because $F$ is probabilistic. Thus, it is at least possible that there is a better solution — for instance, by reducing the variances of $P_0$ and $P_1$, or by moving their means symmetrically away from the origin, we may be able to preserve $\mathcal{H}_2(w_0) = \mathcal{H}_2(1/2) = 1$ while reducing $\mathcal{H}(F(x)|x))$. This is indeed the case: starting from the stated initial parameter values, 53 steps of gradient descent on the training posterior loss (see below for a discussion of the algorithmic issues arising in finding a local minimum of the posterior loss) results in the solution

$$\mu_0 = -2.140, \sigma_0 = 1.256, \mu_1 = 2.129, \sigma_1 = 1.233 \quad (47)$$

at which point the gradients with respect to all four parameters are smaller than 0.03 in absolute value. This solution has an expected posterior loss of 2.55, as opposed to 2.64 for the initial conditions. Of course, the KL divergence of $(1/2)P_0 + (1/2)P_1$ to the sampling density has increased from the initial conditions.

What algorithm should one use in order to minimize the expected posterior loss on a sample? Here it seems worth commenting on the algebraic similarity between Equation (45) and the iterative minimization performed by EM. In (unweighted) EM, if we have a current solution $(1/2)P_0 + (1/2)P_1$, and sample data $S$, then our next solution is $(1/2)P_0' + (1/2)P_1'$, where $P_0'$ and $P_1'$ minimize

$$-\sum_{x \in S} \left(\frac{P_0(x)}{P_0(x) + P_1(x)} \log(P_0'(x)) \right.$$
$$\left. +\frac{P_1(x)}{P_0(x) + P_1(x)} \log(P_1'(x)) \right). \quad (48)$$

While the summand in Equation (48) and the right-hand-side of Equation (45) appear quite similar, there is a crucial difference. In Equation (48) there is a *decoupling* between the posterior prefactors $P_b(x)/(P_0(x) + P_1(x))$ and the log-losses $-\log(P_b'(x))$: our current guesses $P_b$ *fix* the posterior prefactors for each $x$, and then we minimize the resulting weighted log-losses $-\log(P_b'(x))$ with respect to the $P_b'$, giving our next guess. In Equation (45), no such decoupling is present: in order to evaluate a potential solution $P_b'$, we must use the log-losses *and* posteriors determined by the $P_b'$. An informal way of explaining the difference is that in EM, we can use our current guess $(P_0, P_1)$ to generate random labels for each $x$ (using the posteriors $P_b(x)/(P_0(x) + P_1(x))$), and then minimize the log-losses of the $x$ together with their labels to get $P_0', P_1'$. For the posterior loss, to evaluate $(P_0', P_1')$ we must generate the labels according to $(P_0', P_1')$ as well. Thus, there is no obvious iterative algorithm to minimize the expected posterior loss. An alternative is to let $\mathcal{P}$ be a smoothly parameterized class of densities, and resort to gradient descent on the parameters of $P_0$ and $P_1$ to minimize the posterior loss.

An even more intriguing difference between the posterior loss and the standard mixture log-loss can be revealed by examining their derivatives. Let us fix two densities $P_0$ and $P_1$ over $X$, and a point $x \in X$. If we think of $P_0$ and $P_1$ as representing the mixture $(1/2)P_0 + (1/2)P_1$, and we define $L_{log} = -\log((1/2)P_0(x) + (1/2)P_1(x))$ to be the mixture log-loss on $x$, then

$$\frac{\partial L_{log}}{\partial P_0(x)} = \frac{1}{\ln(2)} \frac{-1}{P_0(x) + P_1(x)}. \quad (49)$$

This derivative has the expected behavior. First, it is always negative, meaning that the mixture log-loss on $x$ is always decreased by increasing $P_0(x)$, as this will give more weight to $x$ under the mixture as well. Second, as $P_0(x) + P_1(x) \to 0$, the derivative goes to $-\infty$.

In contrast, if we define the posterior loss on $x$

$$\begin{aligned}
L_{post} &= -\frac{P_0(x)}{P_0(x) + P_1(x)} \log P_0(x) \\
&\quad -\frac{P_1(x)}{P_0(x) + P_1(x)} \log P_1(x) \quad (50)
\end{aligned}$$

then we obtain

$$\frac{\partial L_{post}}{\partial P_0(x)}$$
$$= \frac{1}{P_0(x) + P_1(x)} \Bigg[ -\log P_0(x)$$
$$+ \frac{P_0(x)}{P_0(x) + P_1(x)} \log P_0(x)$$
$$+ \frac{P_1(x)}{P_0(x) + P_1(x)} \log P_1(x) - \frac{1}{\ln(2)} \Bigg]. \quad (51)$$

This derivative shows further curious differences between the mixture log-loss and the posterior loss. Notice that since $1/(P_0(x) + P_1(x)) \geq 0$, the sign of the derivative is determined by the bracketed expression in Equation (51). If we define $R_0(x) = P_0(x)/(P_0(x) + P_1(x))$, then this bracketed expression can be rewritten as

$$(1 - R_0(x)) \log \frac{1 - R_0(x)}{R_0(x)} - \frac{1}{\ln(2)} \quad (52)$$

which is a function of $R_0(x)$ only. Figure 8 shows a plot of the expression in Equation (52), with the value of $R_0(x)$ as the horizontal axis. From the plot we see that $\partial L_{post}/\partial P_0(x)$ can actually be *positive* — namely, the point $x$ can exhibit a *repulsive* force on $P_0$. This occurs when the ratio $R_0(x) = P_0(x)/(P_0(x) + P_1(x))$ falls below a certain critical value (approximately 0.218). The explanation for this phenomenon is straightforward once we have Equation (8): as long as $P_0$ models $x$ somewhat poorly (that is, gives it small probability), it is preferable that $x$ be modeled as poorly as possibly by $P_0$, so as to make the assignment of $x$ to $P_1$ as deterministic as possible. It is interesting to note that clustering algorithms in which data points have explicit repulsive effects on distant centroids have been proposed in the literature on $K$-means and self-organizing maps [5].

From the preceding discussion, it might be natural to expect that, as for $K$-means, minimizing the posterior loss over a density class $\mathcal{P}$ would be more likely to lead to $P_0$ and $P_1$ that are "different" from one another than, say, classical density estimation over $\mathcal{P}$. This intuition derives from the fact that $P_0$ and $P_1$ repel each other in the sense given above. As for $K$-means, this can be shown in a fairly general manner (details omitted).